\documentclass{article}

\usepackage[nonatbib, preprint]{neurips_2025}

\usepackage[utf8]{inputenc} 
\usepackage[T1]{fontenc}    
\usepackage{hyperref}       
\usepackage{url}            
\usepackage{booktabs}       
\usepackage{amsfonts}       
\usepackage{nicefrac}       
\usepackage{microtype}      
\usepackage{xcolor}         

\usepackage{array}
\usepackage{biblatex}
\usepackage{graphicx}
\usepackage{wrapfig}
\title{Grounding Intelligence in Movement}
\author{
  Melanie Segado \\
  University of Pennsylvania\\
  \texttt{melanie.segado@gmail.com} \\
  \And
  Felipe Parodi \\
  University of Pennsylvania \\
  \And
  Jordan K. Matelsky \\
  University of Pennsylvania \\
  \And
  Michael L. Platt \\
  University of Pennsylvania \\
  \And
  Eva B. Dyer \\
  University of Pennsylvania \\
  \And
  Konrad P. Kording \\
  University of Pennsylvania \\
  \texttt{kording@seas.upenn.edu} \\
}

\begin{document}

\maketitle

\begin{abstract}
 
Recent advances in machine learning have dramatically improved our ability to model language, vision, and other high-dimensional data, yet they continue to struggle with one of the most fundamental aspects of biological systems: movement. Across neuroscience, medicine, robotics, and ethology, movement is essential for interpreting behavior, predicting intent, and enabling interaction. Despite its core significance in our intelligence, movement is often treated as an afterthought rather than as a rich and structured modality in its own right. This reflects a deeper fragmentation in how movement data is collected and modeled, often constrained by task-specific goals and domain-specific assumptions. But movement is not domain-bound. It reflects shared physical constraints, conserved morphological structures, and purposeful dynamics that cut across species and settings. \textbf{We argue that movement should be treated as a primary modeling target for AI.} It is inherently structured and grounded in embodiment and physics.  This structure, often allowing for compact, lower-dimensional representations (e.g., pose), makes it more interpretable and computationally tractable to model than raw, high-dimensional sensory inputs. Developing models that can learn from and generalize across diverse movement data will not only advance core capabilities in generative modeling and control, but also create a shared foundation for understanding behavior across biological and artificial systems. Movement is not just an outcome, it is a window into how intelligent systems engage with the world.
\end{abstract}

\section{Introduction}

Embodied intelligence is fundamentally rooted in movement. Across species and contexts, animals – including humans – demonstrate remarkable adaptability in their interactions with objects and conspecifics, often achieving complex motor behaviors with zero or few-shot learning. Nearly all animals use subtle movements of the face and body to signal intent and convey internal state. All processing of information in the brain, from vision to language, has only one goal: emitting better movement primitives, or tokens, that improve evolutionary fitness.

Biological movements are highly context dependent, as are their interpretations, so a useful model clearly needs to accommodate context inputs. Biological organisms excel at making inferences based on the in-context movements of others – recognizing actions, predicting trajectories, detecting disorders. As such, it is clear that proper biological movement understanding is possible. There is therefore every reason to believe that AI systems should be able to excel at this task as well.

The fact that current AI does not do well with biological movement can be seen as a special case of Moravec's paradox. The paradox tells us that these ‘easy’ motor tasks for animals are precisely the ones AI struggles with most. Interestingly, it's not just producing good movements that AI finds challenging (e.g., for robotics), but all of movement modeling and analysis is challenging, including video generation, and especially in contexts like medicine\ref{OMM-goals-table}. 

Movement, unlike language and vision (or more recent foundation model targets like time series\cite{liang2024foundation, talukder2024totem, goswami2024moment}, graphs\cite{liu2023towards, xia2024anygraph, sun2025riemanngfm}, particles\cite{birk2024omnijet}, and neural data\cite{azabou2023unified, cui2024neuro}), does not have any single prototypical data type to model. Instead, there are many types of data relating to biological movement. This includes visual observations (videos), information about embodiment (physical form, body structure, biomechanical context), and trajectory-based timeseries (e.g., accelerometer data, computer-vision-derived poses). There are even neural and physiological signals recorded during movement (EMG, electrophysiology, EEG). While the surface-level data (pixels, accelerometer readings, neural signals) have varying levels of abstraction and dimensionality, they are often just different views or consequences of the same underlying movements. A good model of movement should be able to effectively use all these modalities.

\begin{wrapfigure}{r}{0.5\textwidth}
\centering
\vspace{-9mm}
  \includegraphics[width=\linewidth]{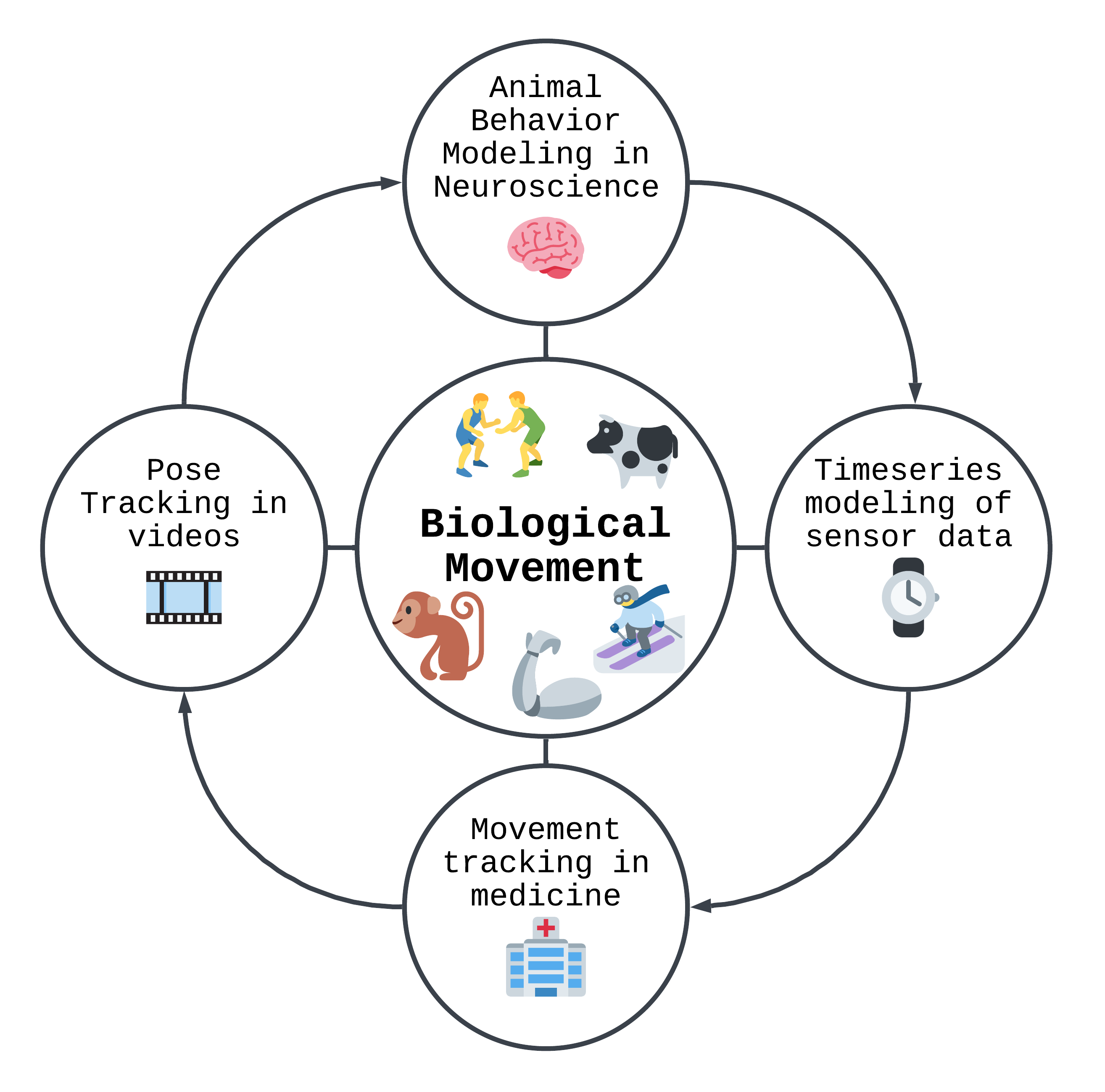}
  \caption{\footnotesize\textbf{Domains with biological movement at their core.} Biological movement is central to neuroscience, medicine, computer vision, and sensor modeling—each offering unique but interconnected perspectives on how movement is tracked, modeled, and understood.}
  \vspace{-3mm}
\end{wrapfigure}

Biological movement is a truly vexing problem: it lies at a unique intersection between highly structured data and high dimensional unstructured data. The body itself is extremely structured, with rough skeletal geometry conserved over hundreds of millions of years \cite{coates1996devonian}. The skeleton and musculature within a species is highly, if not perfectly, conserved \cite{diogo2010muscles}. At the same time, the body around the skeleton is deformable as is the world that animals interact with. As such, biological movement lies in a uniquely interesting part of the world modeling space. 

Meanwhile, existing multi-modal foundation models, such as those focusing on video, fail to model these causal and physical attributes despite having been trained on millions of videos of movement (e.g., producing videos with 180-degree rotations of the neck, or bodies floating up off the ground after a fall). The problem is so hard that extensive tests with current video generative models suggest that anything involving collisions between bodies leads to model failure. Current state of the art models fail catastrophically at generating simple high-fives or stumbling over a rock in our informal experiments. 

Based on an analysis of the movement field, \textbf{our position is that the machine-learning community should intentionally build overarching models of movement, rather than hoping scaled video or task-specific systems will converge on the same capability}.



\section{The landscape of existing movement models}

There are many models through which movement is \textit{partially} represented, such as video generators or time-series modeling\cite{ha2018world, feng2024chatpose, tan2025gaitdynamics}. Collectively, these models are promising as foundational elements for an ML-driven, overarching movement models, since each offers unique and complementary information. Consequently, a unifying framework must effectively integrate elements from these diverse approaches to achieve a more comprehensive understanding of movement.

\subsection{Movement characterization \& understanding}

\paragraph {Pose estimation} Precisely estimating pose is central to movement modeling. Historically this was only possible using marker-based motion capture, which was expensive and not feasible at scale or in natural settings\cite{desmarais2021review, seethapathi2019movement}. However, computer vision has completely revolutionized the field, making it possible to get precise pose estimates across species from hand-held video. Whereas earlier computer-vision approaches often required extensive, task-specific annotated datasets (e.g., for training tools like initial versions of DeepLabCut\cite{mathis2018deeplabcut}, or SLEAP\cite{pereira2022sleap}), foundation model approaches generalize to new species and contexts with minimal (or no) finetuning (e.g., SuperAnimal~\cite{ye2024superanimal}). Human pose estimation specifically has experienced impressive gains with pre-trained models that work across different ages and complex poses, which was previously not possible (e.g., ViTPose \cite{xu2022vitpose, xu2023vitpose++}). This includes the ability to detect and track entities across frames using object-detection models like RT-DETR \cite{zhao2024detrs}. For a long time better pose estimation was a blocker for movement modeling \cite{seethapathi2019movement} but this has effectively been solved by computer vision. 

Bodies are volumetric, and the volumetric properties have a big influence on how they move. Mesh recovery from video is improving and can provide information about body size and geometry (e.g., PromptHMR\cite{wang2025prompthmr}, SMPLer-X\cite{cai2023smpler}, SMAL\cite{zuffi20173d}, Penn Avian Mesh\cite{badger20203d}), hand configurations (MANO\cite{romero2022embodied}), facial expression (DiffusePoseTalk\cite{sun2024diffposetalk}, clothing (ChatGarment\cite{bian2024chatgarment}), and hair (DiffLocks\cite{rosu2025difflocks}). Any model that aims to model movements of specific bodies needs to include all of these features. 

The underlying skeletal structure and musculature of bodies defines \textit{how} they can move. To capture this, the outputs of pose models (2D, 3D, and Mesh) are increasingly being integrated with biomechanical simulators (e.g., OpenSim\cite{delp2007opensim}, Myosuite\cite{caggiano2022myosuite}, Mujoco\cite{todorov2012mujoco}). Models that make explicit use of kinematic data result in estimates that are more aligned with the underlying skeletal structure (e.g., BioPose~\cite{koleini2025biopose}). Biomechanical simulators provide complementary information and are important for training models that unify across representations. 

\paragraph{Action recognition} Actions are composed of sequences of poses, and correctly characterizing these sequences is a key challenge for movement models. Video-action models such as VideoMAE V2\cite{wang2023videomae} already resolve at least 700 distinct human actions\cite{carreira2019short}, yet their animal-centric peers still run an order of magnitude lower: MammalNet\cite{chen2023mammalnet} tracks 12 canonical behaviors across 173 mammal taxa, while ChimpACT\cite{ma2023chimpact} pushes toward finer primate granularity but remains well under the 100-class mark—even when the footage is fully “in-the-wild.” Video foundation models like VideoMAE V2 increasingly achieve robust multimodal understanding by learning shared or aligned representations across modalities (e.g., text, audio). Likewise, multimodal large language models like LLaVA\cite{lin2023video} are increasingly being paired with video and used for action recognition (e.g., LLaVAction~\cite{ye2025llavaction}). 

Beyond video and text, sensor-based activity recognition aims to detect specific activities from wearable sensors such as Inertial Measurement Units (IMUs) or electromyography (EMG) signals. Doing so is critical for applications in human-computer interaction (e.g., for gesture-based wearable controllers \cite{ctrl2024generic} or adaptive prosthetics\cite{parajuli2019real, prasanth2021wearable}) and healthcare (e.g., rehabilitation \cite{wei2023application}. Many models have been proposed for analyzing sensor data, either in isolation or in conjunction with video, including approaches that perform cross-modal prediction from video to sensor timeseries (e.g., \cite{chiquier2023muscles}, and benchmarks like emg2pose~\cite{salter2024emg2pose}). The importance of activity recognition extends beyond human applications. Animal activity recognition, leveraging both video and sensor data is increasingly vital for ecological studies \cite{mao2022fedaar, otsuka2024exploring}, conservation efforts, and applications such as livestock management in agriculture\cite{kleanthous2022deep}.

Ultimately, movement models will need to accommodate and integrate information from many different data streams, likely by combining merits of the growing number of multimodal frameworks (e.g., Act-ChatGPT\cite{nakamizo2024act}, HumanOmni\cite{zhao2025humanomni}, HIS-GPT\cite{zhao2025his}, Meta-transformer\cite{zhang2023meta}, Audiopalm\cite{rubenstein2023audiopalm}, Imagebind\cite{girdhar2023imagebind}, Omnibind\cite{wang2024omnibind}, CoMP\cite{chen2025comp}, M³GPT\cite{luo2024m}) to achieve increasingly nuanced understanding. 

\paragraph{Limitations} Pose estimators  and action recognition models may be able to \textit{describe} movements, but do not typically capture the underlying intent, the nuances of execution quality in a way meaningful for performance analysis or pathology, or the critical environmental and interaction context that gives movement its full meaning. They often struggle with the variability and unscripted nature of real-world actions, particularly those that are not easily categorized or segmented. Furthermore, current approaches often lack the layered understanding required to interpret the same pose sequence differently based on context, such as distinguishing a voluntary handshake from a tremor with diagnostic significance. This highlights the gap between merely describing movement kinematics or labeling actions and truly understanding the dynamic, contextual, and functional aspects of biological movement.

\subsection{Generating movement}

\textbf{Generative Models } The capacity of a model to generate novel, plausible movement serves as a potent indicator of its underlying representational quality. One influential line of research approaches movement generation analogously to language modeling, capitalizing on the compositional structure inherent in poses and actions. Models such as ChatPose \cite{feng2024chatpose} and MotionGPT\cite{jiang2023motiongpt} exemplify this by discretizing continuous movements into a vocabulary of motion "tokens." These are then synthesized using autoregressive strategies, akin to Generative Pre-trained Transformers (GPTs). A key advantage of this token-based paradigm is its natural alignment with textual descriptors, facilitating intuitive text-to-motion generation and enabling tasks like motion infilling (completing sequences between given tokens) and short-term forecasting (predicting subsequent tokens)\cite{jiang2023motiongpt}. 

A movement model should be able to predict what a human/animal's next movement will be, and where they will be at a future point in time. This is an exceptionally complex task, because movement is stochastic and depends on a variety of factors ranging from internal motivations to environmental drivers to physical constraints\cite{pezzulo2024generating}. Models that predict solely based on past movements inevitably fail after very few time-steps \cite{azabou2023relax, orhan2024combining}. To accurately forecast movement, models need to account for goals and intent \cite{diomataris2024wandr, rudenko2020human}.

Moreover, diffusion models\cite{ho2020denoising} are increasingly applied to directly denoise text prompts into structured skeleton-based motion; the Human Motion Diffusion Model (MDM), for example, treats a 3D pose sequence as a trajectory to be denoised step-by-step into fluid, text-conditioned skeletons\cite{tevet2022human,han2024motion}. Building on this, MotionDiffuse \cite{zhang2024motiondiffuse} incorporates foot-contact and diversity losses to produce physically plausible skeletal clips from free-form textual prompts. While these successes across both visual rendering and skeletal animation highlight the immense potential for high-fidelity movement generation, extending these capabilities to achieve nuanced, controllable, and holistic full-body articulated motion consistently presents an ongoing objective, suggesting a need to bridge current specialized methodologies.


\paragraph{Physics-based movement simulation} Movements produce and respond to forces in the environment, so generated movements should do so as well. Progress on physics-informed movement generation has been enabled by biomechanical simulators,  the curation of human motion datasets that include motion dynamics (torques and forces \cite{werling2024addbiomechanics}), and also methods like force estimation from volumetric mesh. Generative models trained on motion dynamics data have been shown to replicate accurate human gait kinematics\cite{tan2025gaitdynamics}, and models that integrate a force loss generate more realistic meshes \cite{yuan2023physdiff} showing that the integration of physics-grounded movement data improves real-world relevance.  

\paragraph{Limitations} Biological movement is extremely precise – in the case of some activities the difference between successful and unsuccessful movements on the order of millimeters and fractions of degrees in the spatial domain, and milliseconds in the temporal domain. Auto-regressive motion generators produce very smoothed, average representations of what specific movements \textit{look like}, but fail to produce detailed actions, and will often produce unrealistic or incoherent results when filling in motions or forecasting. Mesh timeseries generation, as it currently stands, is not useful for applications that require precision.  

Generative video remains imprecise, often misrepresenting bodily movements in ways that violate physical plausibility– even when models incorporate physics-based constraints. Recent systems such as Sora \cite{brooks2024video} or Google Veo 2 (now Veo 3) continue to struggle with producing physically realistic depictions of even simple actions, such as stumbling or coordinated interactions like playing the violin. Despite advances in visual fidelity, these models demonstrate a persistent gap in understanding the structured, biomechanical nature of movement.

\subsection{Learning to move in world models}

\paragraph{Reinforcement Learning (RL) and world models}
RL offers a powerful framework for training agents to perform complex movement tasks through interaction with an environment and the use of reward functions. In RL, agents learn to optimize their actions to achieve specific goals, such as navigating a terrain or manipulating an object. 

Complementary to RL, world models (e.g., NVIDIA Cosmos\cite{agarwal2025cosmos}, Google Genie 2\cite{parker2024genie}) provide a way for agents to learn a model of their environment's dynamics, enabling them to plan and make decisions more effectively \cite{ha2018world, matsuo2022deep, hafner2023mastering}. By learning to predict the consequences of their actions, agents can use world models to anticipate future states and select optimal movement strategies. 

RL agents trained in world models exhibit a wide range of complex and adaptive behaviors \cite{wu2024ivideogpt}, especially focused on how agents move in relation to their environments \cite{simos2025reinforcement, xiao2025robot}. For instance, world models have enabled the development of sophisticated controllers for legged robots (e.g., Boston Dynamics' Spot), allowing them to achieve agile and robust locomotion over challenging terrains\cite{riener2023robots, lai2024world}. Similarly, RL and world models are being explored to create robots that can interact more naturally and effectively with humans, adapting their movements to social cues and user preferences\cite{cheng2024bio}.

\paragraph{Limitations}
Despite their potential, RL and world models face several limitations. One significant challenge lies in specifying precise goals, as RL fundamentally relies on well-defined goals and reward functions, which can be difficult to specify for the nuanced and multifaceted objectives inherent in complex, real-world movement tasks. Furthermore, the model accuracy of world models is crucial for effective planning, but learning a sufficient model of the environment is often intractable, and inaccuracies can lead to suboptimal or even dangerous behavior (see also: "sim-to-real gap", \cite{vafa2024evaluating, yu2024natural, matsuo2022deep}). Learning realistic constraints on movement also remains a substantial hurdle despite the integration of RL with musculoskeletal models\cite{schumacher2022dep}.

Furthermore, the resulting RL policies, even when learned within sophisticated world models, often fail to capture or transfer knowledge to inherently different movement contexts, such as distinct developmental stages or other species. For instance, an RL agent trained to emulate adult human locomotion will not inherently understand or replicate an infant's fidgeting, thereby limiting its use in developmental modeling even though the two motions are deeply biologically linked. Similarly, the movement strategies learned by such an agent are typically not informed by, nor easily adaptable to, the diverse locomotion patterns of other animal species without extensive, separate retraining. Traditionally, this retraining involves a complete redefinition of the agent's morphology and reward structures within the RL framework.~\cite{ao2023curriculum} This task-specific nature means that the movement understanding derived from many RL and world model systems remains highly specialized, posing a hurdle for creating models with truly general or foundational insights into movement.

\begin{figure}
  \centering
  \includegraphics[width=\textwidth]{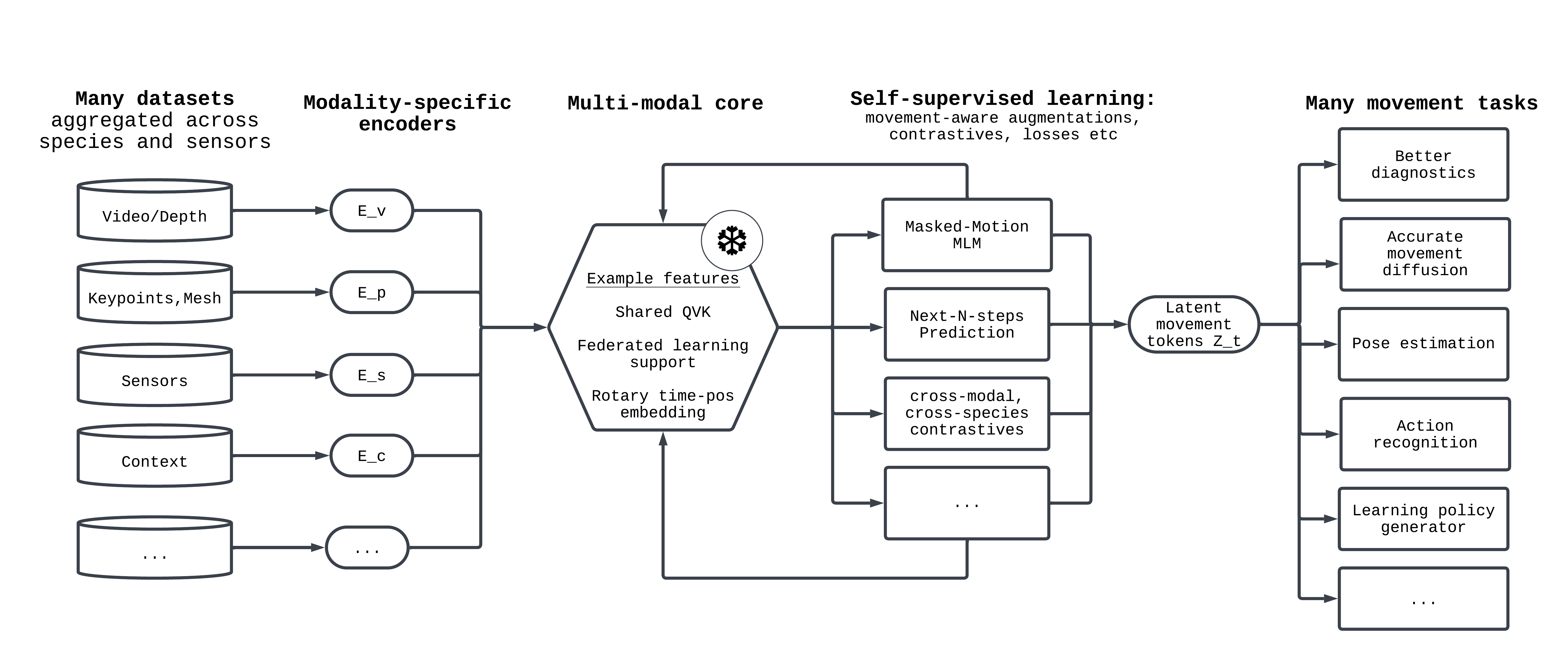}
  \caption{\footnotesize\textbf{Unifying framework for movement modeling across species and sensors.} The ML community has already developed all of these components. What's needed is coordinated effort to combine them into a purpose-built framework for learning movement features directly from aggregated movement data.}
  \label{fig: omm}
\end{figure}

\section{Towards a unifying framework for movement modeling}

Despite impressive improvements in the past few years, existing approaches do not model movement as a unified concept, nor are they precise enough for real-world use. We propose that it is therefore necessary to model movement itself  using a unifying framework, not just as a product of video or sensor data, but as a structured, embodied, and generalizable domain (\textbf{Fig.~\ref{fig: omm}}). 

The resulting models should satisfy the following: 
\begin{enumerate}
    \item \textbf{Cross-modal}, integrating signals such as video, IMU, EMG, GPS, neural data, and facial movements.
    \item \textbf{Biomechanically and physically constrained} across bodies and species.
    \item \textbf{Context-aware}, capable of grounding motion in tasks, environments, and goals
    \item \textbf{Generalizable} across settings, supporting low-shot adaptation in clinical, ecological, and interactive applications.
    
\end{enumerate}
 
 These are not simply larger action recognizers or more detailed movement generators. They are overarching models that treat movement as a fundamental signal, like language or vision, capable of supporting a wide range of scientific and applied domains. Accomplishing this will require coordinated effort to combine existing datasets and fill in gaps, pre-train using identity/privacy preserving approaches that prioritize movement realism, and evaluate on benchmarks that are tailored to real-world impact.  

Below, we outline the steps that are needed to model movement, and the contributions that the ML community can make to enable it. 

\subsection{Step 1: Aggregate a movement data pile} A first step towards a unifying framework is acknowledging the sheer volume and diversity of existing movement data, spanning multiple species, contexts, and sensor modalities. High-fidelity motion capture (MoCap) systems continue to provide gold-standard 3D kinematic data, with established datasets like AMASS ( \cite{mahmood2019amass}) and Human3.6M\cite{ionescu2013human3} being complemented by newer collections capturing nuanced whole-body kinematics (e.g., \cite{ghorbani2021movi}) and rich human-object interactions (e.g.,\cite{lu2025humoto}). The ubiquity of video recording has led to an explosion of data from both controlled and in-the-wild settings. Efforts like Motion-X (\cite{lin2023motion}) are unifying legacy datasets and incorporating new web-scraped videos to create internet-scale resources of human motion meshes. 

For animal studies, datasets such as MammAlps (\cite{tuia2025mammalps}) leverage camera traps for large-scale wildlife monitoring. Furthermore, egocentric and multimodal capture approaches are providing rich, first-person perspectives; EgoBody\cite{zhang2022egobody} pairs head-mounted RGB-D video with eye-gaze and full-body meshes in social scenes, while Nymeria\cite{ma2024nymeria}) utilizes advanced sensors like Project Aria\cite{engel2023project} alongside third-person cameras for human activity capture in natural environments. Complementing these are extensive logs from low-cost wearable sensors, exemplified by CAPTURE-24 \cite{chan2024capture} with over 2,500 hours of free-living wrist-accelerometer data. Physiological data streams, such as the HD-sEMG "Hyser" bank featuring EMG data for hand gestures\cite{jiang2021open}, add another layer of detail. Domain-specific databases are also being contributed, from industrial datasets monitoring worker movements to agricultural datasets like MmCows\cite{vu2024mmcows}, which records synchronized UWB tags, inertial sensors, climate logs, and extensive video footage of dairy cattle. This vast and varied "data pile" underscores the opportunity and challenge for developing comprehensive movement models.

\textbf{Open challenges the ML community can address:}

\textit{Standardizing conventions}: We have lots of data, but combining it is extremely complicated. Developing standardized dataset conventions (similar to the BIDS format for neuroimaging data\cite{appelhoff2019mne}), and building data-loaders that can accept flexible input types and translate into a standard convention (e.g., building on efforts like OpenMMLab's keypoint converter, or HumanML3D\cite{guo2022generating}), would enable aggregation across datasets/sensors/species. 

\textit{Curating datasets to fill in gaps}: Movement datasets need to include more context. This includes pairing video/sensor data with audio, text-based action labels, medical scores/diagnoses, object-interactions, social relationships, time of day, ambient temperature, and physiological measures. While a growing number of multimodal datasets exist, there are still gaps that need to be filled. A priority here is datasets that include both movement data (from one or more sensor streams) and rich, annotated context. 

\subsection{Step 2: Pretrain a multimodal backbone} A flexible, general purpose backbone for movement tasks would be a game-changer for the field of movement science, reducing the time and effort needed to obtain latent features of movement that are useful for downstream tasks. However, for this to have impact, training objectives need to be aligned with the needs of end users. Context needs to be considered, fine-grained details must be preserved, and biomechanical/physical constraints must be strictly respected. 

\textbf{Open challenges the ML community can address:}

\textit{Design architectures for multimodal contextual integration.} Contextual data needs to be included, but the model should not attempt to learn everything about the world – in fact, world models already attempt this – but rather integrate contextual data streams (e.g., environmental sounds, object identities) as 'co-teachers' (akin to approaches in models like CLIP). This will allow them to inform the learned movement latents without having the model attempt to learn everything about the 'world', thereby duplicating efforts in building comprehensive world models. 

\textit{Make federated learning easy.} Much of the interesting, medical movement data that exists can only be analyzed on hospital servers (e.g., videos of newborn infants in the NICU). Federated learning is currently complicated and inaccessible. There is a huge opportunity for the ML community to build user-friendly, open source federated learning training pipelines so models can be trained on sensitive medical data. 

\textit{Develop movement-aware augmentations} Standard image augmentations degrade critical features in biological movement data. Noise corrupts subtle signs like tremor, and left–right flips erase lateralized deficits. Domain-informed augmentations are essential, and the ML community should work closely with experts to formalize and encode their expertise.

\textit{Design movement-aware loss functions}: Realism should be prioritized. This should include adopting losses for joint-angle violations, non-physical accelerations, bone-length inconsistencies, implausible forces, and contact violations to improve realism of generated model outputs\cite{yuan2023physdiff}. 

\subsection{Step 3: Evaluate on high-impact use-cases} 
The true measure of movement models' success should be their practical utility and alignment with societal benefit, especially given the significant resources required for their training. Evaluation must encompass a broader set of critical metrics. This includes assessing generalization and transferability through cross-species and cross-task benchmarks, as well as validating biomechanical and physical realism using metrics explicitly geared to diagnostic needs for movement disorders.  

Model performance evaluation should also consider privacy robustness, ensuring reliable performance even when employing privacy-preserving methods.

\textbf{Open challenges the ML community can address:}

\textit{Develop outcome-aligned benchmarks.} 
In many domains, the utility of a movement model is binary: either it captures the core features of movement — or it fails. Near-miss representations are diagnostically meaningless. Evaluation must reflect this and reward models that produce functionally valid latent representations. Specifically, benchmarks could test causal understanding through physical perturbation recovery or counterfactual movement generation under anatomical or pathological constraints. 

\textit{Benchmark cross-domain generalization.} 
Movement models should be tested across species, body plans, and tasks. Can a model trained on infant reaching also represent gait in Parkinson’s patients or agility in non-human animals? Cross-domain generalization is key to scalability. Models must demonstrate they can predict compensatory patterns, preserve diagnostic asymmetries, and correlate with clinical outcomes – not merely reproduce statistical regularities in training data.

\textit{Incentivize robustness to privacy constraints.} 
Evaluation protocols should reward models that perform well under privacy preserving methods, ensuring that real-world deployment in healthcare and research is feasible and ethical.

\section{Societal stakes \& Ethical risks}

Movement models must be built with explicit attention to risks to privacy, algorithmic bias, and personal safety. Movement data—whether drawn from medical settings, wearables, social behavior, or animal tracking—poses unique risks. It is inherently biometric, often longitudinal, and deeply entangled with questions of identity, ability, health, and physical safety. Modeling it at scale, with generative and predictive capability would be incredibly beneficial—but also opens the door to misuse. These concerns are echoed by many in the AI community who have voiced increasing unease about the centralization of power, the opacity of learned representations, and the potential for social harm when AI systems are developed without robust guardrails. Without appropriate safeguards, movement models could be co-opted for surveillance, behavioral scoring, or automated exclusion, thereby undermining the very communities they are intended to help.

\paragraph{Risks to privacy} Movement data, by its nature, often contains identifiable biometric information such as gait patterns, posture, and even subtle facial expressions captured in video. Recent work has shown that diffusion and transformer-based generative models are prone to memorization of training data, including identifiable faces, gestures, and sequences \cite{gu2023memorization}. In the context of movement modeling, this raises concerns of biometric re-identification, even when datasets are nominally anonymized.

\paragraph{Bias, algorithmic phrenology} The misuse of movement data to infer sensitive traits like health conditions or psychological states carries a significant risk of bias and discrimination in areas such as employment, insurance, and surveillance. Furthermore, models trained on narrowly defined datasets risk performing algorithmic phrenology, reviving discredited attempts to link physical traits to the nonphysical. There is also a critical concern that movement models could unintentionally marginalize individuals with disabilities by defining ‘normal’ or ‘typical’ movements in a way that excludes or misinterprets the movements of those with different physical abilities.

\paragraph{Risk mitigation} To mitigate these risks, we advocate for a design approach rooted in responsible AI development principles. This means building privacy-respecting pipelines (e.g., federated or split learning), embedding morphological and demographic variation during training, and ensuring transparency around model intent, capabilities, and limitations. It also means engaging with domain experts and impacted communities throughout the model development process, not after deployment.

Concerns about potential harms underscore the importance of intentional design. Movement models have the potential to improve clinical care, conservation, assistive technology, and performance science—but only if they are aligned with values of equity, safety, and scientific rigor. By embedding the lessons of AI safety from the outset, we can shape this new modeling paradigm toward outcomes that benefit society as a whole.


\section{Alternative views}

\paragraph{Scaling existing models will solve this}

We are proposing that purpose-built movement models are necessary to advance the field, but one could argue that same progress in modeling movement could simply be achieved by continuously scaling existing, specialized models. This perspective suggests that the increasing realism and familiarity of current approaches—from generative video models that treat movement as pixels to simulators that require ground-up environment construction, or RL agents that demand extensive goal specifications—could eventually converge onto something like a overarching movement model if simply scaled further. The increasing visual fidelity and physical realism  achieved by these scaled systems might make this an appealing proposition.

However, we argue that scaling existing approaches will not solve the fundamental challenges inherent in truly understanding and generating sophisticated biological movement. Scaling alone fails to address core issues such as physical realism, interpretability, and embodiment unless curated constraints are applied. Critically, these models, even at larger scales, will struggle to tackle complex and interesting questions unique to movement science, such as cross-species comparisons and translations, developmental changes, and the subtle manifestations of neurological disorders. 

Furthermore, current modeling paradigms are largely species- and modality-specific, meaning they would not seamlessly scale to include diverse medical or animal data as a unified whole unless such integration is intentionally designed. Duplicating specialized pipelines across various domains is not only costly and inefficient but also precludes the emergence of generalizable movement intelligence. A dedicated, unified framework for movement modeling, which intentionally learns fundamental properties of movement across all data types and species, is essential to move beyond the limitations of scaled, fragmented approaches.

\paragraph{Task specific models are better}

The assertion that a movement model trained on all movement data across tasks and species will outperform task-specific models derives from the massive success of foundation models in language and vision. Admittedly, earlier attempts to create broad "generalist" models (e.g., GATO\cite{reed2022generalist}, Unified IO\cite{lu2022unified}) yielded mixed results, often performing on par with, or sometimes worse than, existing specialized models.

This will not be the case for an overarching movement model. Unlike previous generalist models which were trained on disparate tasks often lacking a cohesive underlying principle (evidenced by the lack of transfer between skills like image captioning and RL policy making), a dedicated movement model would focus on a coherent set of tasks all reliant on precise movement representations. Such a model is therefore well-positioned to benefit from increasingly diverse training data, leveraging strong indications that proficiency in one movement task frequently enhances performance in others. Conceptually, this approach aligns more closely with highly successful architectures like GPT-4\cite{achiam2023gpt}, Gemini\cite{team2023gemini}, and Llama\cite{touvron2023llama}, which excel by deeply modeling specific, rich domains, rather than with the earlier, less potent generalist AIs.

Beyond achieving better performance, a main advantage to training a flexible movement 'backbone' would be to streamline development efforts and increase adoption outside the field of ML. The fragmented nature of movement has resulted in widespread duplication of effort, particularly when applied to real-world problems. Each researcher, doctor, or engineer has to weave together a patchwork of different libraries, each requiring their own packages and data-formatting pipeline, many of which are never supported beyond the conference they were submitted to. Hundreds of new models with incremental advances are published each year, making it near-impossible for researchers outside the field to even know which models are available to them. While initiatives like OpenMMLab\cite{contributors2021openmmlab} have significantly lowered the barrier to entry, the workflow still stands in stark contrast to the ease with which non-ML scientists can now analyze and generate text.


\section{Conclusion}

Movement modeling matters. Developing a model that can recognize, understand, forecast, and generate movement would transform our understanding of movement across species, transform our ability to recognize early signs of disorder and develop precision rehabilitation strategies, and advance the development of robots that predict and synchronize with our movements. 

Here, we argued that overarching models of movement should be developed as a distinct and essential focus for ML systems, and present a framework to unify across species and sensors. We argue that existing approaches have only approximated some elements of movement, and that none have truly modeled movement as a unified concept. The widespread success of task-general foundation models in various other domains, and the growing momentum to develop movement models across all domains, makes this both an ambitious goal but also a logical next step. 

There are massive amounts of movement data that have yet to be aggregated, there exist flexible architectures that can handle diverse data inputs, and there exists scalable compute on which to train. Every component that is needed to model movement exists – all that is needed is coordinated effort across domains. 


The machine learning community is uniquely positioned to take on this challenge, recognizing movement as a foundational domain with the potential for broad scientific impact, clinical breakthroughs, and real-world applications. By uniting efforts across disciplines and data sources, we can build models that not only reflect how organisms move, but also how they act, adapt, and learn. Embracing movement as a core modality opens the door to a deeper understanding of intelligence, understood not only as an abstract computational process but as something grounded in physical interaction, goal-directed behavior, and adaptation to a dynamic world.

{\printbibliography}

\newpage
\appendix

\section*{Technical Appendices and Supplementary Material}

\begin{table}[h]
  \caption{Examples of challenging, high-impact goals in movement modeling and analysis}
  \label{OMM-goals-table}
  \centering
  \renewcommand{\arraystretch}{1.2}  
    \resizebox{0.97\columnwidth}{!}{%
  \begin{tabular}{>{\raggedright\arraybackslash}p{3cm} >{\raggedright\arraybackslash}p{5.5cm} >{\raggedright\arraybackslash}p{6cm}}
    \toprule
    \textbf{Domain} & \textbf{Example goal} & \textbf{Challenging task} \\
    \midrule
    \multicolumn{3}{l}{\textbf{Neuroscience}} \\
    Neuromotor interfaces & High-dimensional control & Intent prediction and action recognition \\
    Neuroethology & Understand social dynamics & Relating movements with neural activity during natural behavior\\
    \midrule
    \multicolumn{3}{l}{\textbf{Health \& Medicine}} \\
    Rehabilitation & Personalize rehabilitation plans and outcome tracking using movement data & Low-shot, context-aware adaptation across patient populations \\
    Biomarker Discovery & Detect early-stage motor degradation in Parkinson’s & Fine-grained temporal modeling that preserves diagnostic signals \\
    Child Development & Diagnose atypical motor development in infants within first few months of age & Learn from rare, unlabeled or unstructured motor trajectories \\
    Climate \& Environmental Health & Track exertion, instability, or gait adaptation in extreme heat or smoke & Movement models conditioned on environmental covariates \\
    \midrule
    \multicolumn{3}{l}{\textbf{Human--Machine Interaction}} \\
    Assistive Devices & Optimize prosthetics or exoskeletons in daily use & Dynamic adaptation to user-specific control patterns \\
    Human--Robot Collaboration & Anticipate and respond to human movement intent & Shared control spaces for fluid joint action \\
    \midrule
    \multicolumn{3}{l}{\textbf{Science}} \\
    Evolutionary Biology & Compare locomotion across phylogenetic lineages & Latent spaces that reflect morphological and functional divergence \\
    Animal Behavior \& Conservation & Model behavioral adaptation across habitats & Transfer across species, climates, and movement vocabularies \\
    \midrule
    \multicolumn{3}{l}{\textbf{Performance, Adaptation, and Resilience}} \\
    Peak Performance & Optimize motion in high-stakes sports and occupational tasks & Detection of micro-asymmetries and fatigue under real-world stress \\
    Space Medicine & Track neuromotor adaptation in microgravity & Few-shot generalization to off-Earth conditions, low-signal environments \\
    Extreme Environments & Model movement under cold, hypoxia, or heavy gear & Environment-aware prediction of failure points and adaptation strategies \\
    \midrule
    \multicolumn{3}{l}{\textbf{Creative Domains}} \\
    Generative video & Generate precise movements that accurately reflect text prompts & Physically grounded movement across species \\
    Choreography & Generate novel, physically valid movement sequences & Constrained generative movement models for creativity and rehearsal \\
    \midrule
    \multicolumn{3}{l}{\textbf{Industry \& Public Policy}} \\
    Occupational Safety & Forecast injury risk from asymmetrical or repetitive motion & Real-time risk estimation in safety-critical domains \\
    Public Infrastructure & Design inclusive environments for mobility and access & Simulation of population-level movement from diverse bodies and devices \\
    \midrule
    \multicolumn{3}{l}{\textbf{Robotics}} \\
    Social Robots & Generate human-like movements that allow for social interaction & Perform real-time interpretation of human movement and generate appropriate adaptive responses  \\
    Legged Robots & Match the ease with which animals adapt to different terrains & Implementation of biomimetic strategies \\
    \bottomrule
  \end{tabular}
  }
\end{table}

\end{document}